\documentclass{article}

\PassOptionsToPackage{numbers, compress}{natbib}
\usepackage[preprint,nonatbib]{neurips_2024}

\usepackage[utf8]{inputenc} % allow utf-8 input
\usepackage[T1]{fontenc}    % use 8-bit T1 fonts
\usepackage[pagebackref, breaklinks=true, colorlinks,citecolor=citecolor, linkcolor=linkcolor, bookmarks=false]{hyperref}
\usepackage{url}            % simple URL typesetting
\usepackage{booktabs}       % professional-quality tables
\usepackage{amsfonts}       % blackboard math symbols
\usepackage{nicefrac}       % compact symbols for 1/2, etc.
\usepackage{microtype}      % microtypography
\usepackage[table]{xcolor}
\usepackage{tablefootnote}
\usepackage{array}       % 扩展表格功能
\usepackage{arydshln} 
\usepackage{wrapfig} % 确保在导言区已加载
\usepackage[export]{adjustbox}
\usepackage{caption,adjustbox}

\newcommand{\our}{{SAMP}\xspace}
\definecolor{myorange}{RGB}{244,133,0}
\definecolor{myblue}{RGB}{30,144,255}
\definecolor{mygreen}{RGB}{46,139,87}

\definecolor{mygreen1}{RGB}{18,253,2}
\definecolor{myred1}{RGB}{255,0,3}
\definecolor{myblue1}{RGB}{0,240,255}
\usepackage{amsmath,amsfonts,bm}
\usepackage{pifont}

\def\eqref#1{(\ref{#1})}
\def\1{\bm{1}}

\DeclareMathAlphabet{\mathsfit}{\encodingdefault}{\sfdefault}{m}{sl}
\SetMathAlphabet{\mathsfit}{bold}{\encodingdefault}{\sfdefault}{bx}{n}

\usepackage{graphicx}
\usepackage{url}            % simple URL typesetting
\usepackage{booktabs}       % professional-quality tables
\usepackage{nicefrac}       % compact symbols for 1/2, etc.
\usepackage{microtype}      % microtypography
\usepackage{wrapfig}

\usepackage{enumitem}
\usepackage[ruled,noend,linesnumbered]{algorithm2e}

\usepackage{multirow}
\usepackage{tablefootnote}

\usepackage{color}
\usepackage{colortbl}
\usepackage{xcolor}

\definecolor{blgrey}{rgb}{0.6,0.6,0.6}
\definecolor{bblue}{rgb}{0.855,0.933,0.98}
\definecolor{dblue}{HTML}{5297D6}
\definecolor{gainred}{rgb}{0.1,0.5,0.3}
\definecolor{citecolor}{HTML}{0071BC}
\definecolor{linkcolor}{HTML}{ED1C24}

\usepackage{listings}
\usepackage{color}

\definecolor{dkcyan}{cmyk}{1,0,0,.25}
\definecolor{dkgreen}{rgb}{0,0.6,0}
\definecolor{gray}{rgb}{0.5,0.5,0.5}
\definecolor{mauve}{rgb}{0.58,0,0.82}

\newcommand{\thickhline}{\noalign{\hrule height 1pt}}

\newcommand{\bs}[1]{\boldsymbol{#1}}

\newcommand{\reals}{\mathbb{R}}

\vspace{-2pt}
\title{SMAP \raisebox{-0.15\baselineskip}{\includegraphics[scale=0.125]{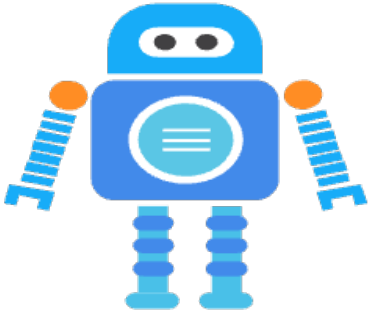}}: Self-supervised Motion Adaptation for Physically Plausible Humanoid Whole-body Control}

\vspace{-5pt}
\author {
    \textbf{Haoyu Zhao}\textsuperscript{* \rm 1, \rm 2},
    \textbf{Sixu Lin}\textsuperscript{* \rm 3},
    \textbf{Qingwei Ben}\textsuperscript{\rm 2, \rm 4},
    \textbf{Minyue Dai}\textsuperscript{\rm 5},
    \textbf{Hao Fei}\textsuperscript{\rm 6},\\
    \textbf{Jingbo Wang}\textsuperscript{\rm 2},
    \textbf{Hua Zou}\textsuperscript{\textdagger \rm 1},
    \textbf{Junting Dong}\textsuperscript{\textdagger \rm 2}\\
    \textsuperscript{1}School of Computer Science, Wuhan University \quad
    \textsuperscript{2} Shanghai Artificial Intelligence Laboratory \quad \\
    \textsuperscript{3} Harbin Institute of Technology (Shenzhen) \quad
    \textsuperscript{4} The Chinese University of Hong Kong \quad \\
    \textsuperscript{5} Fudan University \quad
    \textsuperscript{6} National University of Singapore \quad \\
}

\usepackage{amsthm}

\begin{document}

\maketitle
{
\renewcommand{\thefootnote}{\fnsymbol{footnote}}
\footnotetext{* Equal contributions.}
\footnotetext{\textdagger Corresponding Author.}
}

\begin{figure}[ht]
\vspace{-16pt}
\begin{center}
	\includegraphics[width=\linewidth]{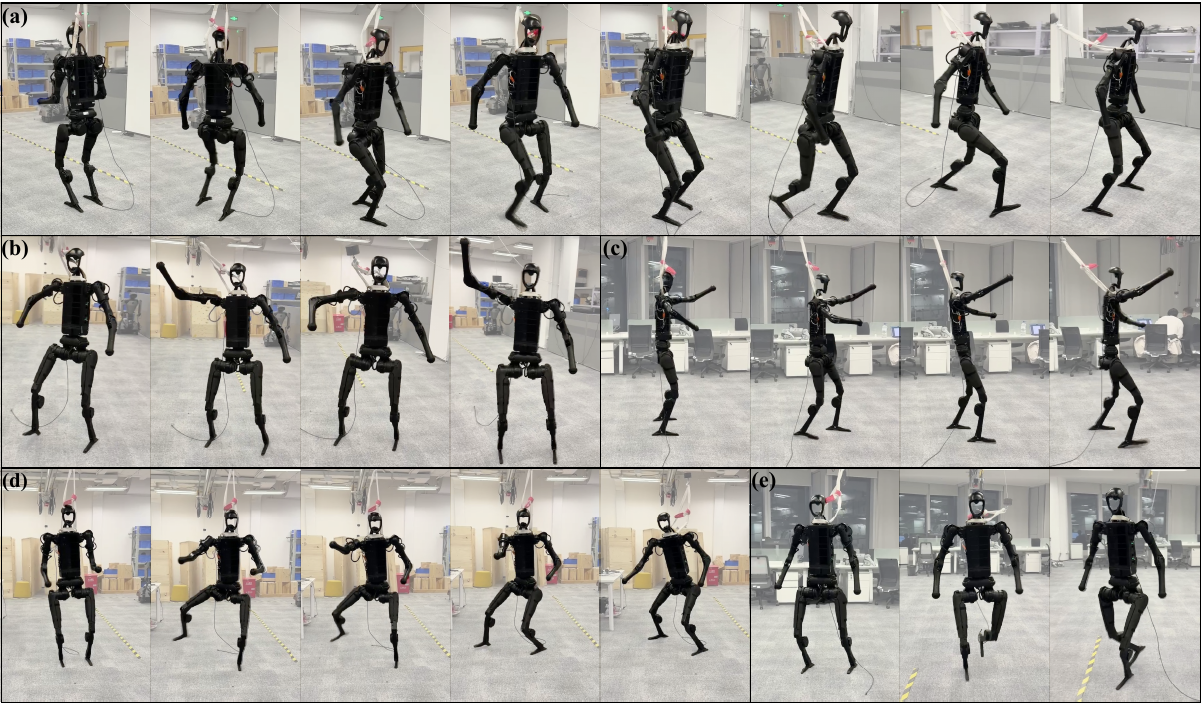}
\end{center}
\vspace{-7pt}
\caption{Our framework enables humanoid robot execute various expressive whole-body motions. The robot can (a) turn around and walk forward, (b) wave hello, (c) swing arms while advancing, (d) jump on one leg, (e) walk fast.}
\vspace{-2pt}
\label{fig:abs}
\end{figure}

\begin{abstract}
\vspace{-4pt}
This paper presents a novel framework that enables real-world humanoid robots to maintain stability while performing human-like motion. Current methods train a policy which allows humanoid robots to follow human body using the massive retargeted human data via reinforcement learning. 
However, due to the heterogeneity between human and humanoid robot motion, directly using retargeted human motion reduces training efficiency and stability.
To this end, we introduce \textbf{\our}, a novel whole-body tracking framework that bridges the gap between human and humanoid action spaces, enabling accurate motion mimicry by humanoid robots.
The core idea is to use a vector-quantized periodic autoencoder to capture generic atomic behaviors and adapt human motion into physically plausible humanoid motion. This adaptation accelerates training convergence and improves stability when handling novel or challenging motions.
We then employ a privileged teacher to distill precise mimicry skills into the student policy with a proposed decoupled reward.
We conduct experiments in simulation and real world to demonstrate the superiority stability and performance of \our over SOTA methods, offering practical guidelines for advancing whole-body control in humanoid robots. 
Our project page is at \href{https://smap-project.github.io/}{https://smap-project.github.io/}.
\end{abstract}

\vspace{-2pt}
\section{Introduction} \label{sec:intro}
\vspace{-4pt}
Humanoid robots, with their human-like morphology, have long been a focal point in robotics due to their potential to perform diverse daily tasks~\cite{zhao2024automated,li2025teleopbench}. Designed for human environments, tools, and interactions, human-sized humanoids serve as ideal platforms for general-purpose robotics, naturally adapting to tasks suited for humans.
% However, realizing this versatility demands precise and robust whole-body control, as humanoid robots must coordinate high-degree-of-freedom movements to interact safely and effectively with their surroundings. require full-body coordination, such as object manipulation, locomotion, and human-robot collaboration.
However, achieving this versatility requires precise and robust whole-body control, enabling humanoid robots to coordinate high-degree-of-freedom movements for safe and effective interaction with their surroundings.
% Unlike wheeled or quadrupedal robots, humanoids must actively maintain dynamic balance while performing tasks that require full-body coordination, such as object manipulation, locomotion, and human-robot collaboration.

% By imitating human behavior, humanoids can harness this rich repertoire of skills and motions, advancing toward general robot intelligence~\cite{zhao2024automated}.

\begin{wrapfigure}{r}{0.45\textwidth}
\centering
% \vspace{-8pt}
\includegraphics[width=\linewidth]{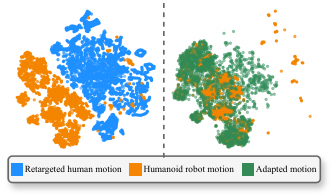}
% \vspace{-12pt}
\caption{\small
\textbf{t-SNE visualization} of the distribution of \textbf{\textcolor{myblue}{retargeted human motion}}, \textbf{\textcolor{myorange}{humanoid robot motion}} (recorded within the simulator), and \textbf{\textcolor{mygreen}{motion adapted by Humanoid-Adapter}} on the CMU MoCap dataset~\cite{cmu_mocap}.}
\label{fig:tsne}
\end{wrapfigure}

Traditional approaches, which decompose the problem into perception, planning, and tracking while modularizing arm and leg control separately~\cite{chestnutt2005footstep,feng2014optimization,kuindersma2016optimization}, are time-consuming to design, limited in scope. 
% Furthermore, these works are highly dependent on contact measurement~\cite{otani2017adaptive}, leading to reliance on expensive setups such as the exoskeleton~\cite{ishiguro2020bilateral}, force sensors, or motion capture systems. 
These limitations make it challenging to scale humanoids to the diverse range of tasks and environments in which they are expected to operate. Human motion capture datasets~\cite{mahmood2019amass,guo2022generating,punnakkal2021babel,cmu_mocap} provide a rich source of reference motion, enabling the imitation of everyday human activities~\cite{zhao2024sg,zhao2024chase}. With the growing availability of large-scale human motion datasets, recent approaches~\cite{fu2024humanplus,cheng2024expressive,he2024omnih2o,he2024learning,ji2024exbody2,lu2024mobile,he2025asap} leverage Reinforcement Learning (RL) to track and mimic retargeted human motion, allowing humanoid robots to learn versatile behaviors. For example, HumanPlus~\cite{fu2024humanplus} presents a system that enables humanoids to learn and imitate human motion and skills in real time using RL. 
% For example, HumanPlus~\cite{fu2024humanplus} presents a system that enables humanoids to learn and imitate human motion and skills in real time using RL. Similarly, Mobile-TeleVision~\cite{lu2024mobile} decouples upper-body control and locomotion, utilizing predictive motion priors and RL to improve manipulation and locomotion robustness. 
However, a major challenge lies in the significant heterogeneity of retargeted human motion data. Given that humanoid robots and humans have entirely distinct action spaces, directly using human motion data as an imitation goal often results in physically implausible motion, leading to low training efficiency and instability. This poses a compelling research question: \textit{How to formulate imitation goals that ensure both physical plausibility and human-like motion for humanoid robots}?

% However, a key challenge lies in the significant heterogeneity of retargeted human motion data. Since humanoid robots and humans operate in fundamentally different action spaces, directly using human motion as an imitation goal often results in physically implausible motions, leading to instability and inefficient training. This raises a critical research question: \textit{How can we define imitation goals that ensure both physical feasibility and human-like motion for humanoid robots}?

To address the aforementioned challenges, we propose \textbf{SMAP}, an effective framework for \textbf{S}elf-supervised \textbf{M}otion \textbf{A}daptation, enabling \textbf{P}hysically plausible whole-body control for humanoid robots. Unlike previous methods that operate within the heterogeneous retargeted human action space, we train and perform inference within the physically plausible action space for humanoid robots. 
Specifically, we introduce \textbf{Humanoid-Adapter}, a vector-quantized periodic autoencoder that maps human motion to humanoid robot actions. 
By encoding human motion sequences into a shared codebook, we decompose motion into generic atomic behaviors and then decoding them into corresponding motion for humanoid robot, our method enables an efficient transformation between human and robot movements, as shown in Fig.~\ref{fig:tsne}. The adapted motion, serving as an imitation goal, significantly improves policy stability and accelerates training convergence. Additionally, through teacher-student distillation and a decoupled reward that separately optimizes upper and lower body dynamics, we further enhance both motion performance and stability.
% The learned policy generates control actions that enable a real humanoid robot to execute the motion in the physical world. 
Extensive experiments on humanoid platforms, Unitree H1, demonstrate that our method achieves superior performance in full-body tracking accuracy and velocity tracking while maintaining stability in dynamic environments. In summary, our work makes the following contributions:
\begin{itemize}
    \item We propose a novel framework for training a robust whole-body control policy that addresses the heterogeneity between human and humanoid action spaces.
    \item We propose a vector-quantized periodic autoencoder that adapts human motion into physically plausible motion for training and inference.
    \item Experiments in simulation and real world validate our method’s superior motion imitation and stability.
\end{itemize}

\vspace{-2pt}
\section{Related Work} \label{sec:related}
\vspace{-2pt}

\subsection{Humanoid Whole-Body Control}
Humanoid robots have great potential to unlock the full capabilities of humanoid systems, but remains a long-standing challenge due to their high degrees of freedom (DoF) and non-linearity~\cite{grizzle2009mabel,hirai1998development}. Traditional approaches often rely on human motion capture suits~\cite{darvish2019whole,dragan2013legibility,ben2025homie},  haptic feedback devices~\cite{brygo2014humanoid,peternel2013learning,ramos2019dynamic}, and dynamics modeling and control~\cite{dariush2008whole,miura1984dynamic,ramos2019dynamic,westervelt2003hybrid,yin2007simbicon}. Recent advances in sim-to-real reinforcement learning (RL) and sim-to-real transfer show promising results in enabling complex whole-body skills for humanoid robots such as walking~\cite{agarwal2023legged,he2024learning,ito2022efficient,cheng2023legs,escontrela2022adversarial,fu2023deep,fuchioka2023opt,yang2023neural,radosavovic2024real,radosavovic2024humanoid}, jumping~\cite{he2025asap}, parkour~\cite{zhuang2024humanoid}, dancing~\cite{fu2024humanplus,cheng2024expressive,he2024omnih2o}, and hopping~\cite{ji2024exbody2}.
% For example, HumanPlus~\cite{fu2024humanplus} presents a system that enables humanoids to learn and imitate human motion and skills in real time using RL. Similarly, Mobile-TeleVision~\cite{lu2024mobile} decouples upper-body control and locomotion, utilizing predictive motion priors and RL to improve manipulation and locomotion robustness. 
For example, H2O~\cite{he2024learning} presents an RL-based teleoperation framework using a third-person RGB camera to capture the human teleoperator’s full-body keypoints. Exbody~\cite{cheng2024expressive} focuses on imitating upper-body reference motion (retargeted from human data) while allowing the legs to robustly follow a given velocity. However, these methods directly use retargeted human motion, which leads to inefficient training and unstable performance due to the significant heterogeneity between human and humanoid robot motion. To address this, we propose Humanoid-Adapter that adapt human motion into the humanoid robot action space, facilitating more efficient and stable learning.

% \begin{figure}
%   \centering
%     \includegraphics[width=\linewidth]{fig/adapter.pdf}\\
%   \vspace{-13pt}
%     % \caption{\textbf{Pipeline}. (\protect\includegraphics[scale=0.02]{fig/frozen.pdf})}
%     % \caption{\textbf{Pipeline}. (\adjustbox{valign=c}{\includegraphics[scale=0.03]{fig/frozen.pdf}})}
%     \caption{\textbf{Humanoid-Adapter}. To align heterogeneous human motion $\mathcal{S}^h$ and humanoid robot motion $\mathcal{S}^r$, we train two VQ-PAEs (\raisebox{-0.4\baselineskip}{\includegraphics[scale=0.018]{fig/train.pdf}}) on them to learn a shared phase manifold using a codebook.}
%     \label{fig:adapter_pipeline}
% \end{figure}

\subsection{Motion Retargeting}
Traditional approaches~\cite{lee1999hierarchical,tak2005physically} often focus on optimizing low-level motion representations or transferring motions to new skeletons, rather than learning a shared representation. This is a common approach used in previous humanoid whole-body control methods~\cite{fu2024humanplus,cheng2024expressive,he2024omnih2o,he2024learning,ji2024exbody2,lu2024mobile,he2025asap}. Recent deep learning advances~\cite{villegas2018neural,aberman2020skeleton,li2023ace} shift towards learning a common cross-character latent space. These methods still require paired data or rely on skeletons for effective learning and auxiliary loss application. For instance, HumanConQuad~\cite{kim2022humanconquad} introduces a human motion-based control interface for quadrupedal robots. Our method achieves a similar effect by leveraging unpaired human motion and RL-generated humanoid robot motion, aligning them to produce physically plausible motion.

\subsection{Motion Manifold Learning}
Motion manifold learning has the primary goal of comprehending the fundamental structures inherent in human movement and dynamics~\cite{li2024walkthedog,starke2022deepphase}. Its distinctive ability to generate human movement patterns presents numerous opportunities to comprehend intrinsic motion dynamics, manage nonlinear relationships in motion data, and acquire contextual and hierarchical representations~\cite{yang2023qpgesture,raab2023modi}. Holden et al.\cite{holden2016deep} generate character movements by mapping high-level parameters to the human motion manifold, enabling diverse motion. Recently, DeepPhase\cite{starke2022deepphase} propose Periodic Autoencoder to learn a low-dimensional motion manifold. With vector quantized periodic autoencoder, we learn a shared phase manifold for human and humanoid robot. The discrete amplitude vectors serve as a narrow bottleneck to regularize unsupervised learning of semantic motion clusters.

\section{Preliminaries} \label{sec:preliminaries}
\medskip
\noindent
\textbf{Goal-conditioned Reinforcement Learning.}
In this work, we train a goal-conditioned policy $\pi$ to achieve whole-body control by imitating reference motion. The learning process is formulated as a Markov Decision Process (MDP), defined by the tuple $\mathcal{M} = \langle \mathcal{S}, \mathcal{A}, \mathcal{T}, \mathcal{R}, \gamma \rangle$, which includes the state space $\mathcal{S}$, action space $\mathcal{A}$, transition dynamics $\mathcal{T}$, reward function $\mathcal{R}$, and discount factor $\gamma$. The state $s_t \in \mathcal{\bs S}$ and transition dynamics $\mathcal{\bs T}$ are determined by the physics simulation~\cite{makoviychuk2021isaac}, which computes the action $a_t$ based on the policy $\pi$. The state $s_t$ consists of proprioception $s_t^p$ and a goal state $s_t^g$.
% Proprioception is defined as $\selfstate \triangleq (\simp, \simv)$, where \(\simp\) represents the 3D body pose, and \(\simv\) denotes velocity. 
The proprioception captures the robot’s internal dynamics, while the goal state $s_t^g$ provides task-specific guidance.
The reward function is defined as $r_t = \mathcal{R}(s_t^p, s_t^g)$, encouraging the policy $\pi$ to maximize long-term rewards. To optimize the policy $\pi$, we employ Proximal Policy Optimization (PPO)\cite{schulman2017proximal}, maximizing the objective 
$\mathbb{E}\left[\sum_{t=1}^{T} \gamma^{t-1} r_{t}\right]$. 

The action space for humanoid robot is represented as $a_t \in \reals^{n \times 3}$, where $n$ is the number of actuated degrees of freedom (DoFs). Each DoF is controlled by a proportional-derivative (PD) controller, and the action $a_t$ specifies the PD target. Our humanoid model follows the kinematic structure of the Unitree H1 robot which has 23 DoFs.

\begin{figure*}[t]
  \centering
    \includegraphics[width=\linewidth]{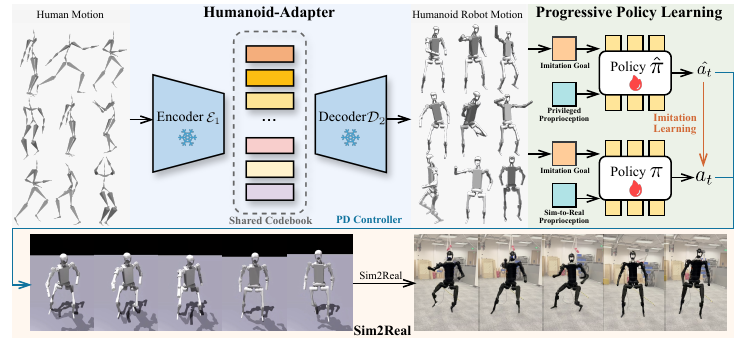}\\
  \vspace{-9pt}
    % \caption{\textbf{Pipeline}. (\protect\includegraphics[scale=0.02]{fig/frozen.pdf})}
    % \caption{\textbf{Pipeline}. (\adjustbox{valign=c}{\includegraphics[scale=0.03]{fig/frozen.pdf}})}
    \caption{\textbf{Pipeline of SMAP \raisebox{-0.2\baselineskip}{\includegraphics[scale=0.08]{fig/logo.pdf}}}. Given human motion, we use the proposed \textbf{Humanoid-Adapter} (details shown in Fig.~\ref{fig:adapter}), pre-trained (\raisebox{-0.5\baselineskip}{\includegraphics[scale=0.02]{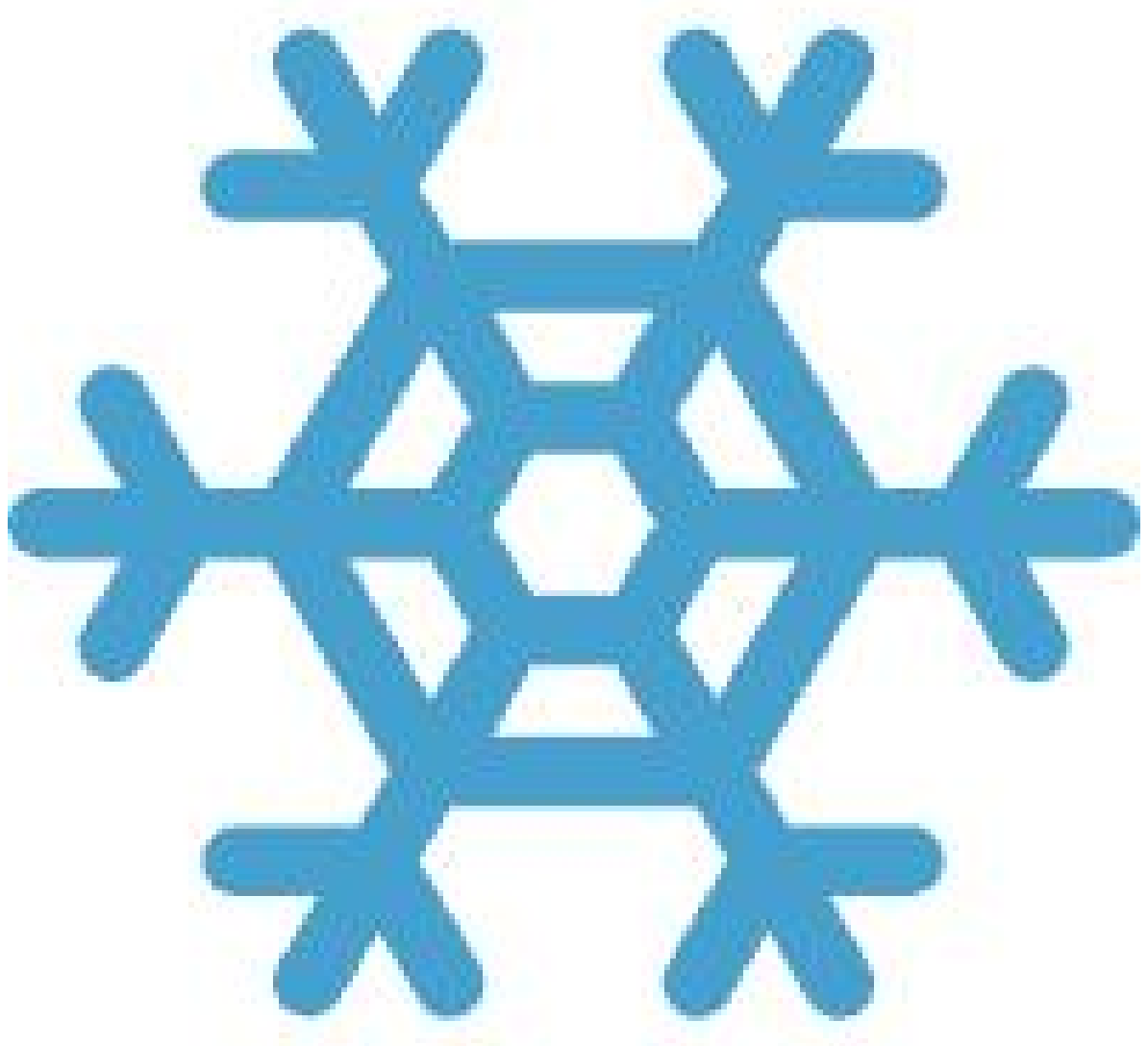}}) to adapt human motion into corresponding, physically plausible humanoid robot motion. Our sim-to-real policy (\raisebox{-0.4\baselineskip}{\includegraphics[scale=0.018]{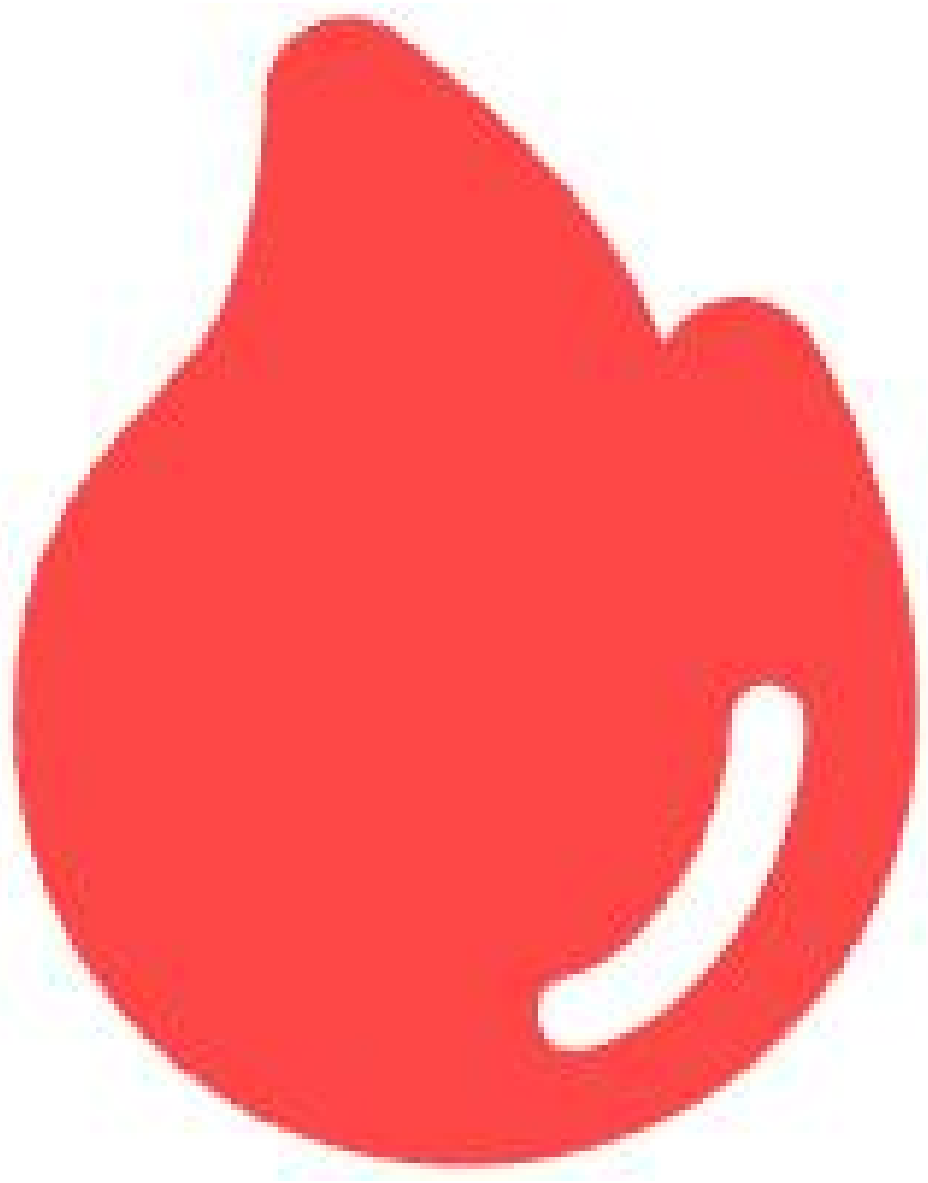}}) is distilled via imitation learning from an RL-trained privileged teacher policy that leverages privileged information with proposed decoupled reward. The policy is transferred to the real world.}
    \label{fig:pipeline}
\end{figure*}

\section{Method} \label{sec:method}
We introduce \textbf{SMAP}, an effective sim-to-real framework for robust whole-body humanoid control, as illustrated in Fig.~\ref{fig:pipeline}. To deal with the the heterogeneity between human and humanoid robot motion, we propose \textbf{Humanoid-Adapter} to adapt human motion action space into physically plausible humanoid robot action space. We then propose \textbf{Progressive Control Policy Learning}, which leverages teacher-student distillation to incorporate privileged inputs and employs a decoupled reward to enhance both the performance and stability of the humanoid robot’s motion.

\subsection{Humanoid-Adapter}
\begin{wrapfigure}{r}{0.55\textwidth}
\centering
% \vspace{-8pt}
\includegraphics[width=\linewidth]{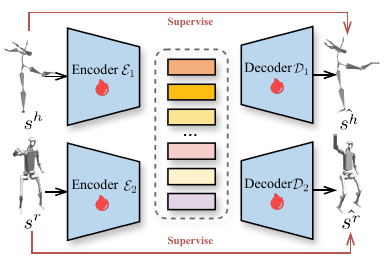}
% \vspace{-20pt}
\caption{\textbf{Humanoid-Adapter}. To align heterogeneous human motion $\mathcal{S}^h$ and humanoid robot motion $\mathcal{S}^r$, we train two VQ-PAEs (\raisebox{-0.4\baselineskip}{\includegraphics[scale=0.018]{fig/train.pdf}}) on them to learn a shared phase manifold using a codebook.}
\label{fig:adapter_pipeline}
\end{wrapfigure}

To transform human motion action space into humanoid robot motion action space, we pre-train Humanoid-Adapter, built upon the Periodic Autoencoder (PAE)~\cite{starke2022deepphase}, to learn a continuous phase manifold, align motion, and cluster semantically similar movements, as shown in Fig.~\ref{fig:adapter_pipeline}. 

Inspired by~\cite{li2024walkthedog,starke2022deepphase}, we aim to learn a generative phase representation for both human and humanoid robot motion, enabling indefinite motion synthesis while preserving temporal coherence and dynamic consistency. To build a humanoid robot motion dataset, $\mathcal{S}^r$, we train a goal-conditioned RL-based policy~\cite{cheng2024expressive} and record the humanoid robot’s motion data within the simulator. Using both the humanoid robot motion dataset $\mathcal{S}^r$ and the human motion dataset~\cite{cmu_mocap} $\mathcal{S}^h$, we learn a shared phase manifold for human and humanoid robot characters without any supervision.

\medskip
\noindent
\textbf{Phase Manifold.}
The Humanoid-Adapter models the latent variational distribution using the phase parameter, which is extracted from the latent motion curve, referred to as the variational phase manifold. This phase manifold is highly structured, capturing key motion characteristics such as timing, local periodicity, and transitions, which are crucial for learning motion features~\cite{li2024walkthedog}. 
% Given an input motion sequence $\textbf{X} \in \mathds{R}^{J \times T}$, where $J$ and $T$ represent the degrees of freedom (DOF) and the number of frames, respectively, our goal is to map each frame $\textbf{X}_i$ to a point $p$ on the phase manifold $\mathcal{P}$:
Given an input motion sequence $d$, our goal is to map each frame $d_i$ to a point $p$ on the phase manifold $\mathcal{P}$:
\begin{equation}
p = \Psi(\alpha, \phi) = \alpha^0\sin(2\pi \phi) + \alpha^1 \cos(2\pi \phi),
\label{eq:phase_manifold1}
\end{equation}
where $\phi$ is the phase parameter, and $\alpha$ is a vector amplitude composed of two components: $\alpha^0$ and $\alpha^1$, representing its first and second halves, respectively. To learn a discrete amplitude space, we employ the vector quantization to cluster the vector amplitude $\alpha$ into a learnable codebook $\mathcal{C}$, represented as:
\begin{equation}
\mathcal{C} = (c_1, c_2, c_3 ... c_n),
\label{eq:phase_manifold2}
\end{equation}
where n is the space size and each $c_i$ is a vector embedding that represents a atomic behavior. Motion with similar characteristics are placed close to each other.
% Notably, there is a one-to-one mapping between each ellipse $\mathcal{P}_i$ and its corresponding amplitude $\alpha_i$. This structure enables flexible scaling of the bottleneck size by adjusting the size of the codebook $\mathcal{A}$.

\medskip
\noindent
\textbf{Structure of Humanoid-Adapter.}
Humanoid-Adapter enables learning a shared phase manifold for both human and humanoid robot motion, without the need for supervision. This is achieved by leveraging the discrete structure, ensuring that semantically similar motion are clustered along the same curve of the manifold. 
% Additionally, motion within each component are temporally aligned through the phase variable, facilitating consistent representation across different motion action space. 

We use an encoder to predict the amplitude $\alpha$, phase $\phi$ and frequency $f$, with vector quantization applied to select the nearest amplitude from a finite codebook $\mathcal{C}$. We then assume the amplitude $\alpha$ and phase $\phi$ hold for the whole input motion sequence and extrapolate the phase linearly with the predicted frequency to the whole sequence. We calculate the embeddings using Eq.~\ref{eq:phase_manifold1} with extrapolated phases $\phi$ and amplitudes $\alpha$. A decoder is then used to reconstruct the input motion sequence $d$ from the predicted embedding. 
% We provide more details of Humanoid-Adapter in our supplementary material.

As shown in Fig.~\ref{fig:adapter_pipeline}, to align human and humanoid robot motion, a phase manifold is learned using a shared codebook $\mathcal{C}$. We train two VQ-PAEs on two datasets: the human motion dataset $\mathcal{S}^h$ and the humanoid robot motion dataset $\mathcal{S}^r$, each with distinct skeletal structures. The loss function for training these networks can be expressed as follows:
\begin{equation}
\mathcal{L} =  \| s^r - \hat{s^r} \|_2 + \| s^h - \hat{s^h} \|_2,
\label{eq:loss}
\end{equation}
where $s^r \in \mathcal{S}^r$, $s^h \in \mathcal{S}^h$, and $\hat{s^r}$, $\hat{s^h}$ are adapted motion. Directly optimizing Eq.~\ref{eq:loss} may cause imbalanced codebook utilization, a common issue in VQ-VAE frameworks due to codebook underutilization. We adapt CVQ-VAE's reinitialization~\cite{zheng2023online} to maintain active embeddings across both motion domains. This method dynamically replaces underused codes during training, ensuring balanced representation learning for both data $S^r$ and $S^h$.

During inference, we use the encoder $\mathcal{E}_1$ and the decoder $\mathcal{D}_2$ to adapt human motion into humanoid robot motion. The proposed Humanoid-Adapter effectively bridges the domain gap between human motion and physically executable humanoid robot motion, ensuring kinematically feasible and dynamically stable motion, as shown in Fig.~\ref{fig:tsne}.

% ====================================================================================================

\subsection{Progressive Control Policy Learning}
During real-world whole-body control of a humanoid robot, much of the information (\textit{e.g.} global linear/angular velocity, positions of each link, and physical properties) which is available in simulation, is not accessible. To address this, our control policy employs a progressive two-stage teacher-student training. In the first stage, the teacher policy is trained using privileged information that can only be obtained in simulation. In the second stage, we replace this privileged information with real-world observations, and distill the teacher policy into a student policy.

\medskip
\noindent
\textbf{Curriculum-based Teacher Policy Training.}
The teacher policy $\hat{\pi}$ takes privileged proprioception and imitation goals as inputs and outputs the action $\hat{a_t}$. The privileged information $s_t^{privileged}$ includes ground-truth states of the humanoid robot and environment (e.g., root velocity, body link positions, and physical properties). For details on privileged information, please refer to the Supp.Mat.

% The training process follows a progressive curriculum strategy, gradually transitioning supervision from reconstructed, physically plausible humanoid robot motion data (generated by the Humanoid-Adapter) to retargeted human motion data. This curriculum learning approach ensures more stable policy convergence compared to training directly on raw retargeted motion. 

The training process follows a progressive curriculum strategy, initially leveraging reconstructed, physically plausible humanoid robot motion data (generated by the Humanoid-Adapter) to stabilize policy learning, while gradually introducing and encouraging exploration of retargeted human motion data. This curriculum learning approach not only ensures convergence stability but also expands the policy's action space.

\medskip
\noindent
\textbf{Student Policy Distilling.}
% In this stage, we remove the privileged information, and use longer history observation to train a student policy. As shown in Figure~\ref{fig:main_figure}, the student policy encodes a series of past observations $o_{t-H:t}$ together with the encoded $g_t$ to get the predicted $a_t \sim \pi(\cdot|o_{t-H:t}, g_t)$. We supervise $\pi$ using the teacher's action $\hat{a}_t \sim \hat{\pi}(\cdot| o_t, g_t)$ with an MSE loss. 
% $$l = \|a_t-\hat{a}_t\|^2$$
% To train the student, we adopt the strategy used in DAgger~\cite{dagger}, we roll out the student policy $\pi$ in the simulation environment to generate training data. For each visited state, the teacher policy $\hat{\pi}$ computes the oracle action as the supervision signal. We proceed to refine the policy $\pi$ by iteratively minimizing the loss $l$ on the accumulated data. The training of $\hat{\pi}$ continues through successive rollouts until the loss $l$ reaches convergence. A critical aspect of training the student policy is preserving a sufficiently long sequence of historical observations. 
In this stage, we remove privileged proprioception and leverage sim-to-real proprioception (a longer history of observations) to train the student policy. 
% The student policy encodes a sequence of past observations along with the encoded goal $s_t^g$ to predict the action $a_t$. 
The policy is supervised using the teacher policy’s action $\hat{a}_t$ with loss followed:
\begin{equation}
\mathcal{L}_{distill} =  \| a_t - \hat{a}_t \|_2.
\label{eq:distill_loss}
\end{equation}

% To train the student, we adopt the strategy from DAgger~\cite{dagger}, rolling out the student policy $\pi$ in the simulation environment to collect training data. For each visited state, the teacher policy $\hat{\pi}$ provides an oracle action as the supervision signal. The student policy $\pi$ is then iteratively refined by minimizing the accumulated loss $\mathcal{L}$. Training continues through successive rollouts until convergence. A crucial factor in effective student policy learning is maintaining a sufficiently long sequence of historical observations to ensure temporal consistency.

To train the student policy, we follow the DAgger~\cite{dagger} framework, rolling out the student policy $\pi$ in the simulation environment to collect training data. At each visited state, the teacher policy $\hat{\pi}$ generates an oracle action as the supervision signal. The student policy $\pi$ is iteratively refined by minimizing the loss $\mathcal{L}_{distill}$.

\begin{table*}[t]
\centering
\small
\setlength{\tabcolsep}{1pt} % 微调列间距
\begin{tabular}{c|cccc|cccc}  % 删除第二列后的列定义 (1 + 4 + 4)
\hline \thickhline
\multirow{2}{*}{\textbf{Method}} & \multicolumn{4}{c|}{$\textbf{Trained Motion Sample}$} & \multicolumn{4}{c}{$\textbf{Novel Motion Sample}$} \\  
\cline{2-9}  % 列范围调整为 2-9
& $E_{vel}\downarrow$ & $E_{mpkpe}\downarrow$ & $E_{mpjpe}\downarrow$ & $fail\downarrow$
& $E_{vel}\downarrow$ & $E_{mpkpe}\downarrow$ & $E_{mpjpe}\downarrow$ & $fail\downarrow$ \\
\hline\hline
Privileged policy & 0.1002 & 0.0531 & 0.0888 & 798 & 0.1923 & 0.0732 & 0.1121  & 199 \\
\hdashline
\rowcolor{gray!10}HumanPlus~\cite{fu2024humanplus} & 0.3103 & 0.1011 & 0.1989 & 3009 & 0.4299 & 0.1598 & 0.2612 & 633 \\
H2O~\cite{he2024learning}  & 0.2333 & 0.0831 & 0.1989 & 2762 & 0.3613 & 0.1385 & 0.2212 & 501 \\
\rowcolor{gray!10}OmniH2O~\cite{he2024omnih2o}  & 0.1791 & 0.0619 & 0.1250 & 1899 & 0.2591 & 0.0912 & 0.1481 & 387 \\
Exbody~\cite{cheng2024expressive}  & 0.2160 & 0.0766  & 0.1783 & 2264 & 0.3002 & 0.1070 & 0.1868 & 432  \\
\rowcolor{gray!10}Exbody\textsuperscript{$ \dagger $}~\cite{cheng2024expressive}  & 0.2285 & 0.0770 & 0.1592 & 2322 & 0.3239 & 0.1099 & 0.1749 & 489 \\
Exbody + AMP~\cite{peng2021amp}  & 0.2499 & 0.0732 & 0.1531 & 1993 & 0.3487 & 0.1002 & 0.1604 & 412  \\
\rowcolor{gray!10} Exbody + Humanoid-Adapter & - & - & - & - & 0.2998 & 0.0999 & 0.1632 & 361 \\
\rowcolor[HTML]{D7F6FF}\textbf{\our} & \textbf{0.1698} & \textbf{0.0608} & \textbf{0.1181} & \textbf{1731} & \textbf{0.2331} & \textbf{0.0893} & \textbf{0.1458} & \textbf{266} \\
\hline\hline
\rowcolor{gray!10} \our w/o Humanoid-Adapter & 0.1743 & 0.0612 & 0.1221 & 1851 & 0.2465 & 0.0921 & 0.1442 & 392 \\
\our w/o teacher-student distillation & 0.2038 & 0.0732 & 0.1521 & 1751 & 0.2765 & 0.0941 & 0.1641 & 389 \\
\rowcolor{gray!10} \our w/o progressive & 0.1712 & 0.0610 & 0.1191 & 1889 & 0.2387 & 0.0914 & 0.1468 & 299 \\
\our w/o decoupled reward & 0.1739 & 0.0659 & 0.1283 & 1775 & 0.2389 & 0.0903 & 0.1499 & 291 \\
\end{tabular}
\vspace{-7pt}
\caption{\textbf{Quantitative Comparisons and Ablation Study}. Simulation-based motion imitation evaluation of our method and state-of-the-art (SOTA) approaches on the CMU MoCap dataset~\cite{cmu_mocap} for the Unitree H1 robot.}
\label{tab:result}
\end{table*}

\medskip
\noindent
\textbf{Decoupled Reward Design} 
% Our reward function is designed to enhance both the performance and stability of the humanoid robot’s motion. It consists of tracking rewards for velocity, direction, orientation of the root, keypoint and joint position. To achieve high-precision control while maintaining whole-body stability, we design tracking rewards that decouple the upper and lower body. Specifically, we assign higher weight on the upper body as it requires greater precision for accurate execution of tasks, while placing lower weight on the lower body to prioritize whole-body balance rather than strict positional accuracy. Additionally, regularization terms are incorporated to improve stability and generalization. The tracking rewards are detailed in Tab.~\ref{tab:rewards_detailed}, while the regularization terms are discussed in the supplementary materials.
Our reward function is designed to improve both performance and stability of the humanoid robot’s motion. It includes tracking rewards for velocity, direction, orientation of the root, keypoint, and joint position. To balance precision and stability, we decouple the upper and lower body: higher weight is assigned to the upper body for precision, while lower weight is given to the lower body to prioritize balance~\cite{ben2025homie}. Regularization terms are also included to enhance stability and generalization. For more detailed tracking rewards, please refer to the Supp.Mat.

\section{Experiments} \label{sec:exp}

\subsection{Implementation Details}
We conduct our experiments in IsaacGym~\cite{makoviychuk2021isaac} simulator. During training, 4096 environments are simulated in parallel on a NVIDIA RTX 4090 GPU.
% NVIDIA RTX 4090 GPU. Detailed hyperparameter settings are available in Appendix A.

\subsection{Dataset}
% \textbf{CMU MoCap}~\cite{cmu_mocap}.
Following Exbody~\cite{cheng2024expressive}, we selectively use a portion of the CMU MoCap dataset~\cite{cmu_mocap}, excluding motion involving physical interactions with other individuals, heavy objects, or rough terrain. This diversity not only enhances the expressiveness of humanoid motion but also improves locomotion stability in unseen scenarios. To further evaluate robustness and generalization, we also test on \textbf{novel motion} from the CMU dataset that are not used during training.

\begin{figure*}[t]
  \centering
    \includegraphics[width=\linewidth]{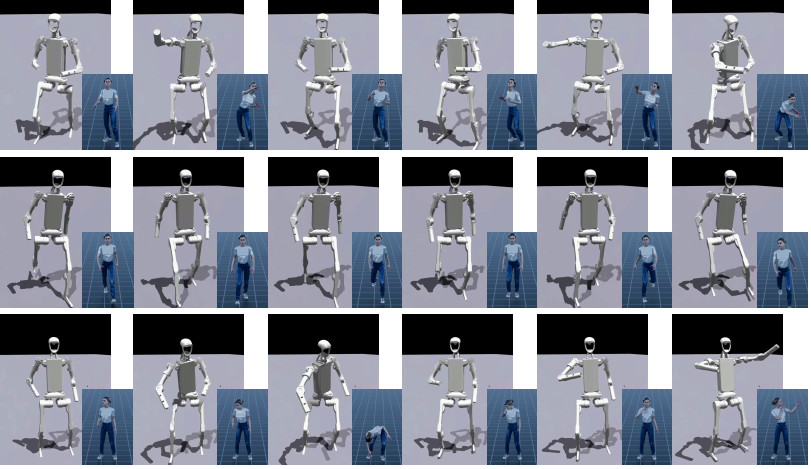}\\
  \vspace{-7pt}
    % \caption{\textbf{Pipeline}. (\protect\includegraphics[scale=0.02]{fig/frozen.pdf})}
    % \caption{\textbf{Pipeline}. (\adjustbox{valign=c}{\includegraphics[scale=0.03]{fig/frozen.pdf}})}
    \caption{\textbf{Qualitative results} on the H1 robot in simulation.}
    \label{fig:result}
\end{figure*}

\subsection{Evaluation Metrics}
We evaluate the policy’s performance using several metrics computed across all motion sequences in the dataset.
\textbf{The mean linear velocity error ($E_{vel}$)} measures the discrepancy between the robot’s root linear velocity and the demonstration, reflecting velocity tracking accuracy.
\textbf{The mean per key point position error ($E_{mpkpe}$)} and \textbf{the mean per joint position error ($E_{mpjpe}$)} evaluate motion accuracy, with $E_{mpkpe}$ assessing keypoint tracking and $E_{mpjpe}$ capturing joint tracking accuracy.
\textbf{Failure ($fail$)} counts the number of failure terminations across all motion sequences. Lower failure rates suggest greater robustness and control consistency across diverse motions.

% \begin{wrapfigure}{r}{0.49\textwidth}
%   \centering
%     \includegraphics[width=\linewidth]{fig/ablation-adapter.pdf}\\
%   \vspace{-15pt}
%     % \caption{\textbf{Pipeline}. (\protect\includegraphics[scale=0.02]{fig/frozen.pdf})}
%     % \caption{\textbf{Pipeline}. (\adjustbox{valign=c}{\includegraphics[scale=0.03]{fig/frozen.pdf}})}
%     \caption{\textbf{Ablation study}. Visualization performance in the simulation on challenging motion sample. }
%     \label{fig:ablation}
% \end{wrapfigure}

% \begin{wrapfigure}{r}{0.55\textwidth}
%   \centering
%     \includegraphics[width=\linewidth]{fig/ablation-progressive.pdf}\\
%   \vspace{-15pt}
%     % \caption{\textbf{Pipeline}. (\protect\includegraphics[scale=0.02]{fig/frozen.pdf})}
%     % \caption{\textbf{Pipeline}. (\adjustbox{valign=c}{\includegraphics[scale=0.03]{fig/frozen.pdf}})}
%     \caption{\textbf{Ablation study} of teacher-student distillation. The \textbf{\textcolor{green}{green}} points represent the imitation goal, while the \textbf{\textcolor{myred1}{red}} points correspond to the DOF position. }
%     \label{fig:ablation_teacher}
% \end{wrapfigure}

\begin{figure}
  \centering
  \begin{minipage}[t]{0.42\textwidth}
    \centering
    \includegraphics[width=\linewidth]{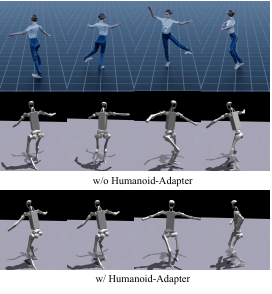}
    \vspace{-15pt}
    \caption{\textbf{Ablation study}. Visualization performance in the simulation on challenging motion sample.}
    \label{fig:ablation}
  \end{minipage}
  \hfill % 添加水平间距
  \begin{minipage}[t]{0.56\textwidth}
    \centering
    \includegraphics[width=\linewidth]{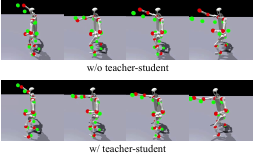}
    \vspace{-15pt}
    \caption{\textbf{Ablation study} of teacher-student distillation. The \textbf{\textcolor{green}{green}} points represent the imitation goal, while the \textbf{\textcolor{myred1}{red}} points correspond to the DOF position.}
    \label{fig:ablation_teacher}
  \end{minipage}
\end{figure}

\subsection{Comparison with SOTA Methods}
We evaluate our method on motion tracking in simulation across Unitree H1, comparing it with three state-of-the-art methods: HumanPlus~\cite{fu2024humanplus}, H2O~\cite{he2024learning}, OmniH2O~\cite{he2024omnih2o}, and Exbody~\cite{cheng2024expressive}. This teacher policy (privileged policy) leverages all privileged environment information as mentioned in its observations. 
For a fair comparison, we reimplement the H2O~\cite{he2024learning} and OmniH2O~\cite{he2024omnih2o} using global keypoint tracking and the same observation space. Since Exbody~\cite{cheng2024expressive} only tracks upper-body motion, we introduce a whole-body tracking variant (Exbody$^\dagger$) to enable direct comparison. 
This version tracks full-body movements based on human motion data. 
Exbody + AMP~\cite{peng2021amp} uses an AMP reward to encourage the policy’s transitions to be similar to those in the retargeted dataset.
% Exbody + AMP~\cite{peng2021amp} utilizes an AMP reward to align the policy’s transitions with those in the retargeted dataset.

% We evaluate on the retargeted CMU MoCap dataset~\cite{cmu_mocap} following Exbody~\cite{cheng2024expressive} in simulation, as shown in Tab.~\ref{tab:result}. Overall, our method significantly improves full-body and velocity tracking accuracy compared to state-of-the-art methods, demonstrating stable and effective performance in dynamic environments. It has the fewest failure cases, highlighting its robustness. Notably, the motion adapted by the proposed Humanoid-Adapter can also enhance the stability of other RL-based methods (Exbody + Humanoid-Adapter), further demonstrating the strong generalizability of our approach.

We evaluate our method on the retargeted CMU MoCap dataset~\cite{cmu_mocap} in simulation, following Exbody~\cite{cheng2024expressive}, as shown in Tab.~\ref{tab:result}. 
% our algorithm performs well in full-body tracking accuracy
% Our algorithm also excels in velocity tracking accuracy, outperforming other methods. This improvement is attributed to the teacher-student training paradigm, where the distillation from privileged information to historical observation results in a student policy with better velocity-tracking.  Additionally, it achieves the fewest failure cases, highlighting its robustness.
% Our method achieves high full-body tracking accuracy and excels in velocity tracking, outperforming other SOTA methods. This improvement stems from the teacher-student training paradigm, where distilling privileged information into historical observations enables the student policy to track velocity more effectively. Additionally, our method has the fewest failure cases, demonstrating its robustness. this is because proposed Humanoid-Adapater. it turn human motion into physically Plausible motion, which is more easy for humanoid robot to tracking as well as keep stable
Our method achieves high full-body tracking accuracy and excels in velocity tracking, outperforming other state-of-the-art methods. This improvement stems from the teacher-student distillation, where distilling privileged information into historical observations enables the student policy to track velocity more effectively. Additionally, our method has the fewest failure cases, demonstrating its robustness. This is largely attributed to the proposed Humanoid-Adapter, which transforms human motion into physically plausible motion, making it easier for the humanoid robot to track while maintaining stability. Notably, \textit{the motion adapted by the proposed Humanoid-Adapter also enhances the stability of other RL-based methods} (Exbody + Humanoid-Adapter), further demonstrating its strong generalizability.

\begin{figure*}[t]
  \centering
    \includegraphics[width=\linewidth]{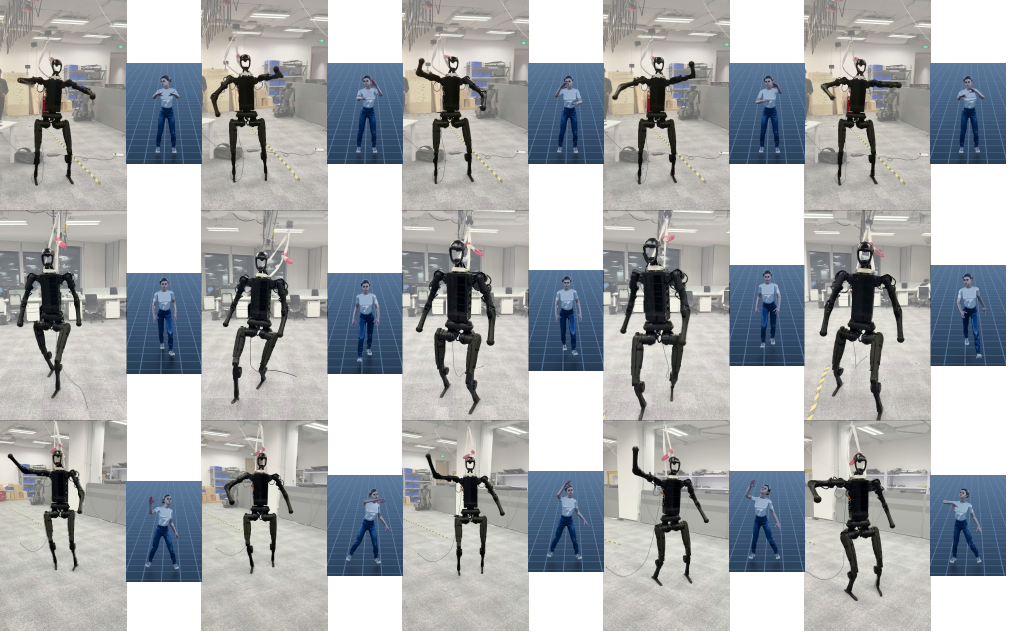}\\
  \vspace{-5pt}
    % \caption{\textbf{Pipeline}. (\protect\includegraphics[scale=0.02]{fig/frozen.pdf})}
    % \caption{\textbf{Pipeline}. (\adjustbox{valign=c}{\includegraphics[scale=0.03]{fig/frozen.pdf}})}
    \caption{\textbf{Qualitative result} on the H1 robot in real world.}
    \label{fig:result_real_world}
\end{figure*}

% Fig.\ref{fig:result} shows the qualitative results in simulation, where the humanoid robot successfully mimics various human motions, such as fast walking and squatting, while maintaining stability. Additionally, Fig.\ref{fig:result_real_world} demonstrates qualitative results in real-world experiments, further validating the robustness of our policy in real-world scenarios. More qualitative results can be found on our \href{https://smap-projectpage.github.io/}{project page}.
Fig.\ref{fig:result} presents qualitative results in simulation, where the humanoid robot successfully replicates various human motions, such as fast walking and squatting, while maintaining stability. Additionally, Fig.\ref{fig:result_real_world} showcases real-world experimental results, further demonstrating the robustness of our policy in real-world settings.

\begin{figure}[t]
  \centering
  \begin{minipage}[t]{0.53\textwidth} % 调整比例关系
    \vspace{0pt} % 强制顶部对齐
    \includegraphics[width=\linewidth]{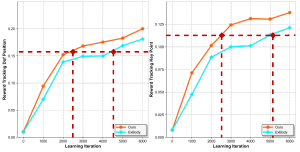}
    \vspace{-12pt} % 图片与标题间距
    \caption{\textbf{Training Curves Comparison} between \textcolor{myorange}{\our} and \textcolor{myblue1}{Exbody\textsuperscript{$\dagger$}}}
    \label{fig:adapter}
  \end{minipage}
  \hfill
  \begin{minipage}[t]{0.42\textwidth}
    \vspace{0pt}
    \centering
    \fontsize{9}{10}\selectfont
     \setlength{\tabcolsep}{4pt}
    \scalebox{0.9}{ % 缩放表格以适应宽度
     \begin{tabular}{ccccc}
    \hline \thickhline
      & \rotatebox{0}{$E_{vel}\downarrow$} & \rotatebox{0}{$E_{mpkpe}\downarrow$} & \rotatebox{0}{$E_{mpjpe}\downarrow$} & \rotatebox{0}{$fail\downarrow$} \\
     \hline \hline
      \multicolumn{5}{@{}c}{\textbf{Codebook Size $\mathcal{C}$}} \\
      \rowcolor{gray!10} 16 & 0.1732 & 0.0611 & 0.1192 & 1798 \\
      32 & \textbf{0.1698} & \textbf{0.0608} & \textbf{0.1181} & \textbf{1731} \\
      \rowcolor{gray!10} 64 & \textbf{0.1698} & 0.0610 & 0.1190 & 1799 \\
      \midrule
      \multicolumn{5}{@{}c}{\textbf{History Length}} \\
      \rowcolor{gray!10} 0 & 0.2831 & 0.0751 & 0.1591 & 1973 \\
      5 & 0.1751 & 0.0623 & 0.1289 & 1787 \\
      \rowcolor{gray!10} 10 & \textbf{0.1698} & \textbf{0.0608} & \textbf{0.1181} & 1731 \\
      32 & 0.1781 & \textbf{0.0608} & 0.1188 & \textbf{1719} \\
    \end{tabular}}
    \vspace{5pt} % 表格与标题间距
    \caption{\textbf{Ablation Study Results} (Best values in bold)}
    \label{tab:ablation1}
  \end{minipage}
\end{figure}

% ===========================================================================================
% ===========================================================================================

% \begin{wraptable}{r}{0.4\textwidth} 
%  \centering
%  \fontsize{9}{10}\selectfont
%  \setlength{\tabcolsep}{3pt}  % 设置列间距
%  \begin{tabular}{ccccc}
% \hline \thickhline
%   & $E_{vel}\downarrow$ & $E_{mpkpe}\downarrow$ & $E_{mpjpe}\downarrow$ & $fail\downarrow$\\ 
%  \hline \hline
%  \multicolumn{5}{c}{\textbf{Size of Codebook $\mathcal{C}$}} \\  
%  \rowcolor{gray!10}16 & 0.1732 & 0.0611 & 0.1192 & 1798\\
%  32 & \textbf{0.1698} & \textbf{0.0608} & \textbf{0.1181} & \textbf{1731}\\
%  \rowcolor{gray!10}64 & \textbf{0.1698} & 0.0610 & 0.1190 & 1799\\
% \hline \hline

% \multicolumn{5}{c}{\textbf{History Length}} \\  
%  \rowcolor{gray!10}0 & 0.2831 & 0.0751 & 0.1591 & 1973\\
%  5 & 0.1751 & 0.0623 & 0.1289 & 1787\\
%  \rowcolor{gray!10}10 & \textbf{0.1698} & \textbf{0.0608} & \textbf{0.1181} & 1731\\
%   32 & 0.1781 & \textbf{0.0608} & 0.1188 & \textbf{1719}\\
%  \end{tabular}
%  \caption{\textbf{Ablation study} on trained motion sample.}
%  \label{tab:ablation1}
% \end{wraptable}

\subsection{Ablation Study}
In this section, we conduct experiments to evaluate the effectiveness of proposed modules in Tab.~\ref{tab:result}

\medskip
\noindent
\textbf{Humanoid-Adapter.} 
To evaluate the impact of Humanoid-Adapter, we replace it with directly retargeted human motion (\our w/o Humanoid-Adapter) during both training and inference. Since retargeted human motion may not always be physically plausible, the tracking performance degrades significantly under novel and challenging motion samples, ultimately reducing robot stability.
Humanoid-Adapter adapts complex motion into more robot-friendly forms, enabling more stable and accurate motion tracking as shown in Fig.~\ref{fig:ablation}.

With more physically plausible motion as the imitation goal, our method also accelerates training convergence, leading to significantly faster improvements in tracking DOF positions and keypoint rewards compared to Exbody$^\dagger$.
Specifically, our approach reaches the same level of performance as Exbody$^\dagger$ in nearly half the iterations—achieving comparable results in just 2500 iterations, whereas Exbody$^\dagger$ requires over 4500 iterations. 

% This demonstrates the efficiency of our method in learning stable and precise motion control.

% With more physically plausible motion as the imitation goal, our method accelerates training convergence, leading to significantly faster improvements in tracking DOF positions and keypoint rewards compared to Exbody$^\dagger$. Specifically, our approach reaches the same level of performance as Exbody$^\dagger$ in nearly half the iterations—achieving comparable results in just 2500 iterations, whereas Exbody$^\dagger$ requires over 4500 iterations. This demonstrates the efficiency of our method in learning stable and precise motion control.

% To assess the impact of the proposed Humanoid-Adapter, we replace it with directly retargeted human motion (\our w/o Humanoid-Adapter) during both training and inference. Since retargeted human motion is not always physically plausible, this leads to significantly degraded tracking performance, especially for novel and challenging motions, ultimately reducing robot stability. As shown in Fig.~\ref{fig:ablation}, Humanoid-Adapter enhances tracking performance on complex motions such as dancing by adapting them into more robot-friendly forms, enabling more stable and accurate motion tracking. Moreover, using physically plausible motion as the imitation goal accelerates training convergence. The improvement in tracking DOF position and keypoint rewards is notably faster compared to Exbody$^\dagger$. After approximately 2500 iterations, our method achieves performance comparable to Exbody$^\dagger$ after more than 4500 iterations.
\medskip
\noindent
\textbf{Progressive Control Policy Learning.} 
We first analyze the impact of teacher-student distillation. Without this training (\our w/o teacher-student distillation), tracking accuracy decreases significantly. This is primarily due to the lack of privileged velocity guidance, which makes it challenging for the single-stage RL policy to learn velocity directly from historical data. As shown in Fig.~\ref{fig:ablation_teacher}, the policy without teacher-student distillation demonstrates lower tracking precision. We then examine the effectiveness of the proposed progressive learning. We find that directly using the final weight for policy training (\our w/o progressive) yields worse results. This is because gradually allowing the model to learn retargeted motion is crucial for improving learning performance.

\medskip
\noindent
\textbf{Decoupled Reward.} 
We observe that the control policy employing the decoupled reward demonstrates significantly higher precision in tracking motion targets and fewer failures. This improvement is due to the upper body’s greater need for precision in task execution, while the lower body is prioritized for overall balance rather than strict positional accuracy.

\medskip
\noindent
\textbf{Hyperparameters of \our.} 
Choosing an appropriate codebook size is critical for our framework, as a small codebook size fails to capture the diverse semantics, and a large codebook reduces semantic alignment accuracy. As shown in Tab.~\ref{tab:ablation1}, a codebook size of 32 yields the best performance.

We also test student policies trained with varying history lengths in Tab.~\ref{tab:ablation1}. Without additional history, the policy struggles to learn effectively. History length of 10 produces the best results, which we use in our experiments. Longer history lengths increase the difficulty of fitting privileged information, ultimately reducing tracking performance.

\vspace{-2pt}

\section{Conclusion}
\label{sec:conclusion}
This paper introduces \textbf{SMAP}, a novel sim-to-real framework for whole-body humanoid control. Different from previous methods, we use a network to map human motion into the humanoid robot’s action space for training and inference. For superior disentanglement, we propose a vector-quantized periodic autoencoder to bridge the gap between human motion and humanoid robots. Then, we propose Progressive Control Policy Learning, leveraging teacher-student distillation and employing a decoupled reward that separately optimizes upper and lower body dynamics. Extensive experiments in simulation and real world demonstrate that our \our achieves superior full-body tracking accuracy while maintaining stability.

\medskip
\noindent
\textbf{Limitation and future work.} 
One limitation of \our is the lack of explicit joint correspondence, which may cause minor mismatches in motion alignment. By sampling motion in action space of humanoid robot, Humanoid-Adapter can be a data augmentation tool to generate reliable motion, offering a promising avenue for future research.

{
    \small
    \bibliographystyle{cite}
    \bibliography{main}
}

\end{document}

\maketitle

%%%%%%%%%%%%%%%%%%%%% NO MORE THAN 14 PAGES %%%%%%%%%%%%%%%%%%%%%

% %%%%%%%%%%%%%%%%%%%%% NO MORE THAN 14 PAGES %%%%%%%%%%%%%%%%%%%%%

\appendix

\section{Real Robot System Setup}
\begin{figure}[h]
  \centering
    \includegraphics[width=0.7\linewidth]{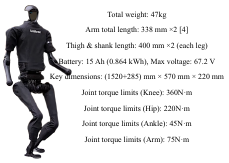}\\
    % \caption{\textbf{Pipeline}. (\protect\includegraphics[scale=0.02]{fig/frozen.pdf})}
    % \caption{\textbf{Pipeline}. (\adjustbox{valign=c}{\includegraphics[scale=0.03]{fig/frozen.pdf}})}
    \caption{\textbf{Details} about Unitree H1 robot. }
    \label{fig:h1}
\end{figure}

Our real robot is built on the Unitree H1 platform, as shown in Fig.~\ref{fig:h1}, equipped with Damiao DM-J4310-2EC motors. The control policy receives motion-tracking target information as input, computes the desired joint positions for each motor, and sends commands to the robot’s low-level interface. The policy’s inference frequency is set at 50 Hz. The commands are sent with a delay kept between 18 and 30 milliseconds. The low-level interface operates at a frequency of 500 Hz, ensuring smooth
real-time control. The communication between the control policy and the low-level interface is realized through LCM
(Lightweight Communications and Marshalling).

\section{More Details of Humanoid Adapter}
We then assume the two properties hold for the whole input sequence $\textbf{X}$ and extrapolate the phase linearly with the predicted frequency to the whole sequence. We calculate the embeddings using Eq.~\ref{eq:phase_manifold}:
\begin{equation}
p = \Psi(\alpha, \phi) = \alpha^0\sin(2\pi \phi) + \alpha^1 \cos(2\pi \phi),
\label{eq:phase_manifold1}
\end{equation}
with extrapolated phases and amplitudes. A decoder is then used to reconstruct the input motion sequence from the predicted embedding. A decent reconstruction can only be achieved if the learned mapping is close to phase linear and amplitude constant.

The encoder uses a 2-layer 1D convolutional network to map the input to an intermediate representation, which is then split into two branches: the timing branch and the amplitude branch. The timing branch predicts phase $\phi$ and frequency $f$ from a temporal signal generated by a 1D convolution, while the amplitude branch predicts amplitude $\textbf{A}$ by applying average pooling followed by an MLP, with vector quantization used to select the nearest amplitude from a finite codebook.

The phase is calculated using the relative timing $\mathcal{T}$ and the equation $\Phi = \phi + f \cdot \mathcal{T}$. The final motion embedding is obtained by $\textbf{P} = \Psi(\textbf{A}, \Phi)$. The decoder is a 2-layer 1D convolutional network that maps the embedding back to the original motion space.

\begin{table}[t]
\centering
\small  % Change font size (use \footnotesize or \scriptsize for even smaller)
\setlength{\tabcolsep}{8pt}
\begin{tabular}{@{}ccc@{}}
\hline \thickhline
\textbf{Term} & \textbf{Expression} & \textbf{Weight} \\ 
\hline
\rowcolor{gray!10}DoF acceleration & $\| \ddot{d}_t \|_2^2$ & $-3e^{-7}$   \\
DoF position limits & $\mathbb{I}(d_t \notin [q_{\min}, q_{\max}])$ & $-10$ \\
\rowcolor{gray!10}DoF error & $\| d_t - d_0 \|_2^2$ & $-0.5$ \\
Energy & $\| \tau_t \dot{d}_t \|_2^2$ & -0.001  \\
\rowcolor{gray!10}Linear velocity (z) & $\| v^{\text{lin-z}}_t \|_2^2$ & -1 \\
Angular velocity (xy) & $\| v^{\text{ang-xy}}_t \|_2^2$ & -0.4  \\
\rowcolor{gray!10}Action rate & $\| a_t - a_{t-1} \|_2^2$ & $-0.1$ \\
Torque & $\| \tau_t \|_2$ & $-0.0001$ \\
\rowcolor{gray!10}Feet air time & $T_{\text{air}} - 0.5$ & $10$ \\
Feet velocity & $\|v_{\text{feet}}\|_1$ & $-0.1$ \\
\rowcolor{gray!10}Feet contact force & $\| F_{\text{feet}} \|_2^2$ & $-0.003$ \\
Stumble & $\mathbb{I}(F_{\text{feet}}^x > 5 \times F_{\text{feet}}^z)$ & $-2$ \\
\rowcolor{gray!10}Hip pos error & $\| p^{\text{hip}}_t - p^{\text{hip}}_0 \|_2^2$ & $-0.2$ \\
Waist roll pitch error & $\| p^{\text{wrp}}_t - p^{\text{wrp}}_0 \|_2^2$ & $-1$ \\
\rowcolor{gray!10}Ankle Action & $\| a^{\text{ankle}}_t \|_2^2$ & $-0.1$ \\
\end{tabular}
\caption{\textbf{Regularization rewards} Regularization rewards for preventing undesired behaviors for sim-to-real transfer.}
\label{tab:rewards_more_detailed}
\end{table}

\begin{table}[t]
\centering
\small
\setlength{\tabcolsep}{5pt} % 适当增加间距提升可读性
\begin{tabular}{@{}lcr@{}} % 优化列对齐
\hline \thickhline
\textbf{Term} & \textbf{Expression} & \textbf{Weight} \\ 
\hline \hline
\rowcolor{gray!10}DoF Position (Upper) & $\exp(-0.7 \|\mathbf{q}_{ref}^{upper} - \mathbf{q}^{upper}\|)$ & 3.0 \\
DoF Position (Lower) & $\exp(-0.7 \|\mathbf{q}_{ref}^{lower} - \mathbf{q}^{lower}\|)$ & 1.0 \\
\rowcolor{gray!10}Keypoint Position (Upper) & $\exp(-\|\mathbf{p}_{ref}^{upper} - \mathbf{p}^{upper}\|)$ & 2.0 \\ 
Keypoint Position (Lower) & $\exp(-\|\mathbf{p}_{ref}^{lower} - \mathbf{p}^{lower}\|)$ & 1.0 \\ 
\rowcolor{gray!10}Linear Velocity & $\exp(-4.0 \|\mathbf{v}_{ref} - \mathbf{v}\|)$ & 6.0 \\
Velocity Direction & $\exp(-4.0 \cos (\mathbf{v}_{ref}, \mathbf{v}))$ & 6.0 \\
\rowcolor{gray!10}Roll \& Pitch & $\exp(-\|\boldsymbol{\Omega}_{ref}^{\phi\theta} - \boldsymbol{\Omega}^{\phi\theta}\|)$ & 1.0 \\
Yaw & $\exp(-|\Delta y|)$ & 1.0 \\
% \hline
\end{tabular}
\caption{\textbf{Tracking Reward.} $\mathbf{q}_{ref}^{upper/lower}$ and $\mathbf{p}_{ref}^{upper/lower}$ denote the reference joint and keypoint positions for the upper and lower body, respectively. $\mathbf{v}_{ref}$ is the reference velocity of the body, while $\boldsymbol{\Omega}_{ref}^{\phi\theta}$ and $\boldsymbol{\Omega}^{\phi\theta}$ denote the reference and actual roll and pitch of the body.}
\label{tab:rewards_detailed}
\end{table}

\section{More Details of Reward}
In the main paper, we introduce the tracking reward. Tab.~\ref{tab:rewards_more_detailed} and Tab.~\ref{tab:rewards_detailed} shows details on the regularization reward.

\section{More Details of Observation of RL}

The student policy's observation comprises the following components: Base angular velocity, IMU measurements (roll and pitch angles), Directional difference between current yaw angle and target yaw (represented as sine and cosine terms), Joint positions and velocities, Observation history from the past n steps, and Target goal.

The teacher policy's observation extends the student policy's observation with privileged information, including: Foot contact flags, System dynamics parameters (mass, ground friction coefficients, motor strength parameters), External push forces.

\begin{table}
\centering
\small
\setlength{\tabcolsep}{9pt} % 适当增加间距提升可读性
\begin{tabular}{lcr} % 移除了 @{} 可能引起冲突
\hline
\hline % 使用标准双线而非 \thickhline
\textbf{Codebook size} & \textbf{E$_{mpkpe}$} \\ % 数学模式下标
\hline
\rowcolor{gray!10}16 & 12.3 \\
32 & 10.6 \\
\rowcolor{gray!10}64 & 10.7 \\
\hline
\end{tabular}
\caption{Mean per joint position error (cm) of Humanoid-Adapter.}
\label{tab:supp_adapter}
\end{table}

\section{Performance of Humanoid-Adapter}
We save a set of paired human motion and humanoid robot motion data (saved by simulator with a RL policy) to evaluate the performance of our Humanoid-Adapter, as shown in Tab.~\ref{tab:supp_adapter}.

\section{Additional Real World Results Visualization}
\begin{figure*}
  \centering
    \includegraphics[width=0.8\linewidth]{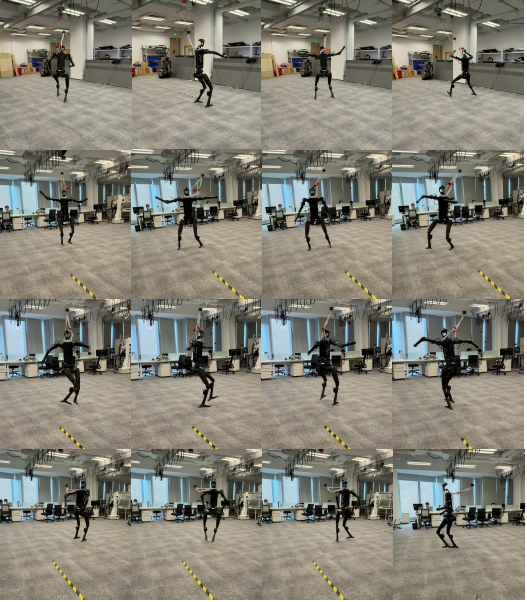}\\
    % \caption{\textbf{Pipeline}. (\protect\includegraphics[scale=0.02]{fig/frozen.pdf})}
    % \caption{\textbf{Pipeline}. (\adjustbox{valign=c}{\includegraphics[scale=0.03]{fig/frozen.pdf}})}
    \caption{ Expressive motion evaluation in the real world..}
    \label{fig:supp}
\end{figure*}

We provide detailed visualization for some motions evaluated in the real world in Fig.~\ref{fig:supp}.

% {
%     \small
%     \bibliographystyle{cite}
%     \bibliography{main}
% }